\documentclass[10pt,twocolumn,letterpaper]{article}

\usepackage{cvpr}
\usepackage{times}
\usepackage{epsfig}
\usepackage{graphicx}
\usepackage{amsmath}
\usepackage{amssymb}

\usepackage{wrapfig}
\usepackage{caption}
\usepackage{subcaption}
\usepackage{color}
\usepackage[dvipsnames]{xcolor} 
\usepackage{xspace}
\usepackage{comment}
\usepackage{paralist}
\usepackage{booktabs}
\usepackage[font={footnotesize},labelfont={footnotesize}]{caption}

\DeclareMathOperator*{\argmax}{arg\,max}

\newcommand{\rama}{\textcolor{black}}

\newcommand{\coco}{COCO}
\newcommand{\justify}{justification}
\newcommand{\disccap}{discriminative image captioning}
\newcommand{\easy}{\emph{easy confusion}}
\newcommand{\hard}{\emph{hard confusion}}
\newcommand{\reffig}[1]{Fig.~\ref{#1}}
\newcommand{\refsec}[1]{Sec.~\ref{#1}}
\newcommand{\reftab}[1]{Table.~\ref{#1}}
\newcommand{\refeq}[1]{Eq.~\ref{#1}}
\newcommand{\cubjust}{CUB-Justify\xspace}

\newcommand{\groundjust}{IS($\lambda$)}
\newcommand{\groundjustmixed}{semi-blind-IS($\lambda$)}
\newcommand{\literaljust}{IS(1)}
\newcommand{\blindjust}{blind-IS($\lambda$)}
\newcommand{\pragma}{RS($\lambda$)}

\newenvironment{packed_itemize}{
\begin{list}{\labelitemi}{\leftmargin=2em}
\vspace{-6pt}
 \setlength{\itemsep}{0pt}
 \setlength{\parskip}{0pt}
 \setlength{\parsep}{0pt}
}{\end{list}}

\usepackage[pagebackref=true,breaklinks=true,letterpaper=true,colorlinks,bookmarks=false]{hyperref}

\cvprfinalcopy 

\def\cvprPaperID{99} 

\ifcvprfinal\fi
\begin{document}

\title{Context-aware Captions from Context-agnostic Supervision}

\author{
	Ramakrishna Vedantam$^{1}$ \quad Samy Bengio$^{2}$ \quad Kevin Murphy$^{2}$ \quad Devi Parikh$^3$ \quad Gal Chechik$^2$\\
	$^1$Virginia Tech \quad $^3$Georgia Institute of Technology \quad $^2$Google\\
	{\tt\small $^1$vrama91@vt.edu} \quad {\tt\small $^3$parikh@gatech.edu} \quad {\tt\small $^{2}$\{bengio,kpmurphy,gal\}@google.com}\\
}

\maketitle

\begin{abstract}
	\rama{We introduce an inference technique to produce discriminative context-aware image captions (captions that describe differences between images or visual concepts) using only generic context-agnostic training data (captions that describe a concept or an image in isolation). For example, given images and captions of ``siamese cat" and ``tiger cat'', we generate language that describes the ``siamese cat'' in a way that distinguishes it from ``tiger cat''. Our key novelty is that we show how to do joint inference over a language model that is context-agnostic and a listener which distinguishes closely-related concepts. We first apply our technique to a justification task, namely to describe why an image contains a particular fine-grained category as opposed to another closely-related category of the CUB-200-2011 dataset. We then study discriminative image captioning to generate language that uniquely refers to one of two semantically-similar images in the \coco{} dataset. Evaluations with discriminative ground truth for justification and human studies for discriminative image captioning reveal that our approach outperforms baseline generative and speaker-listener approaches for discrimination.}
\end{abstract}

\section{Introduction}\label{sec:introduction}

\begin{figure}
	\includegraphics[width=\columnwidth,page=3]{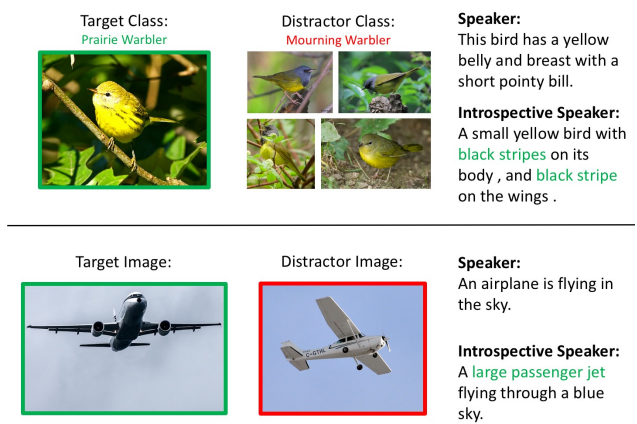} \vspace{-20pt}
	\caption{
    \footnotesize
    An illustration of two tasks requiring pragmatic reasoning explored in this paper. 1) \emph{justification}: Given an image of a bird, a target (ground-truth) class (green), and a distractor class (red), describe the target image to explain why it belongs to the target class, and not the distractor class. The distractor class images are only shown for illustration, and are not provided to the algorithm. 2) \emph{discriminative image captioning}: Given two similar images, produce a sentence to identify a target image (green) from the distractor image (red). Our introspective speaker model improves over a context-free speaker.\vspace{-20pt}}
    \label{fig:teaser}
\end{figure}

Language is the primary modality for communicating, and representing knowledge. To convey relevant information, we often use language in a way that takes into account context. For example, instead of describing a situation in a ``literal" way, one might pragmatically emphasize selected aspects in order to be persuasive, impactful or effective. Consider the target image at the bottom left in~\reffig{fig:teaser}. A literal description ``An airplane is flying in the sky'' conveys the semantics of the image, but would be inadequate if the goal was to disambiguate this image from the distractor image (bottom right). For this purpose, a more pragmatic description would be, ``A large passenger jet flying through a blue sky". This description is aware of context, namely, that the distractor image also has an airplane. People use such pragmatic considerations continuously, and effortlessly in teaching, conversation and discussions.

In this vein, it is desirable to endow machines with pragmatic reasoning. One approach would be to collect training data of language used in context, for example, discriminative ground truth utterances from people describing images in context of other images, or justifications explaining why an image contains a target class as opposed to a distractor class (\reffig{fig:teaser}). Unfortunately, collecting such data has a prohibitive cost, since the space of objects in possible contexts is often too large. Furthermore, in some cases the context in which we wish to be pragmatic may be unknown \emph{apriori}. For example, a free-form conversation agent may have to respond in a context-aware or discriminative fashion depending upon the history of a conversation. Such scenarios also arise in human-robot interaction, as in the case where, a robot may need to reason about which spoon a person is asking for. Thus, in this paper, we focus on deriving pragmatic (context-aware) behavior given access only to generic (context-agnostic) ground truth.

We study two qualitatively different real-world vision tasks that require pragmatic reasoning. The first is \emph{\justify}, where the model needs to justify why an image corresponds to one fine-grained object category, as opposed to a closely related, yet undepicted category. Justification is a task that is important for hobbyists, and domain experts: ornithologists and botanists often need to explain why an image depicts particular species as opposed to a closely-related species. Another potential application for justification is in machine teaching, where an algorithm instructs non-expert humans about new concepts.

Our second task is \emph{\disccap}, where the goal is to generate a sentence that describes an image in context of other semantically similar images. This task is not only grounded in pragmatics, but is also interesting as a scene understanding task to check fine-grained image understanding. It also has potential applications to human robot interaction.

\rama{Recent work by Andreas and Klein~\cite{pragma} derives pragmatic behaviour in neural language models using only context-free data. While we are motivated by similar considerations, the key algorithmic novelty of our work over~\cite{pragma} is a unified inference procedure which leads to more efficient search for discriminative sentences (\refsec{sec:results}). Our approach is based on the realization that one may simply re-use the sampling distribution from the generative model, instead of training a separate model to assess discriminativeness~\cite{pragma}. This also has important implications for practitioners, since one can easily adapt existing context-free captioning models for context-aware captioning without additional training.} Furthermore, while \cite{pragma} was applied to an abstract scenes dataset~\cite{Zitnick_2013_CVPR}, we apply our model to two qualitatively different real-image datasets: the fine-grained birds dataset CUB-200-2011~\cite{WahCUB_200_2011}, and the \coco~\cite{coco} dataset which contains real-life scenes with common objects. 

In summary, the key contributions of this paper are: 
\begin{packed_itemize}
\item A novel inference procedure that models an introspective speaker (IS), allowing a speaker (S) (say a generic image captioning model) to reason about pragmatic behavior without additional training.
\item Two new tasks for studying discriminative behaviour, and pragmatics, grounded in vision: \emph{justification}, and \emph{discriminative image captioning}.
\item A new dataset (\cubjust) to evaluate justification systems on fine-grained bird images with 5 captions for 3161 (image, target class, distractor class) triplets.
\item Our evaluations on \cubjust{}, and human evaluation on \coco{} show that our approach outperforms baseline approaches at inducing discrimination.
\end{packed_itemize}

\section{Related Work}\label{sec:related_work}

\noindent{\textbf{Pragmatics:}} 
The study of pragmatics -- how context influences usage of language, stems from the foundational work of Grice~\cite{grice_1975} who analyzed how cooperative multi-agent linguistic agents could model each others' behavior to achieve a common objective. Consequently, a lot of pragmatics literature has studied higher-level behavior in agents including conversational implicature~\cite{Benotti2009ACA} and the Gricean maxims~\cite{Vogel2013EmergenceOG}. 
These works aim to derive pragmatic behavior given minimal assumptions on individual agents and typically use hand-tuned lexicons and rules. \rama{More recently, there have been exciting developments on applying reinforcement learning (RL) techniques to these problems~\cite{mordatch_emergence_2017,das_learning_2017,lazaridou_multi-agent_2016}, requiring less manual tuning.}

We are also interested in deriving pragmatic behavior, but our focus is on scaling context-sensitive behavior to vision tasks.
Other works model ideas from pragmatics to learn language via games played online~\cite{Wang2016LearningLG} or for human-robot collaboration~\cite{Tellex2014AskingFH}. In a similar spirit, here we are interested in applying ideas from pragmatics to build systems that can provide justifications (\refsec{subsec:cub_justify}) and provide discriminative image captions (\refsec{subsec:setup/disc_image_caption}).

Most relevant to our work is the recent work on deriving pragmatic behavior in abstract scenes made with clipart, by Andreas, and Klein~\cite{pragma}. Unlike their technique, our proposed approach does not require training a second listener model and supports more efficient inference (\refsec{subsec:emitter_suppressor_bs}). More details are provided in~\refsec{subsec:reasoning_speaker}.
\\[2pt]
\noindent{\textbf{Beyond Image Captioning:}} Image captioning, the task of generating natural language description for an image, has seen quick progress~\cite{Donahue_CVPR_2015,Fang_CVPR_2015,Vinyals_CVPR_2015,Xu_ICML_2015}. Recently, research has shifted beyond image captioning, addressing tasks like visual question answering~\cite{VQA,Geman_PNAS_2015,Malinowski_NIPS_2014,Zhu_CVPR_2015}, referring expression generation~\cite{Berg_EMNLP_2014,Mao2015GenerationAC,Nagaraja_ECCV_2016,Sadovnik2013NotES}, and fill-in-the-blanks~\cite{Yu_ICCV_15}. In a similar spirit, the two tasks we introduce here, \emph{justification}, and \emph{discriminative image captioning}, can be viewed as ``beyond image captioning" tasks. Sadovnik~ \etal~\cite{Sadovnik_CVPR_2012} first studied a discriminative image description task, with the goal of distinguishing one image from a set of images. Their approach incorporates cues such as discriminability and saliency, and uses hand-designed rules for constructing sentences. In contrast, we develop inference techniques to induce discriminative behavior in neural models. The reference game from~\cite{pragma} can also be seen as a discriminative image captioning task on abstract scenes made from clipart, while we are interested in the domain of real images. The work on generating referring expressions by Mao~\etal~\cite{Mao2015GenerationAC} generates discriminative captions which refer to particular objects in an image given context-aware supervision. Our work is different in the sense that we address an instance of pragmatic reasoning in the common case where context-dependent data is not available for training.
\\[2pt]
\noindent{\textbf{Rationales:}} Several works have studied how machines can understand human rationales, including enriching classification by asking explanations from humans \cite{Donahue_ICCV_2011}, and 
incorporating human rationales in active learning~\cite{BiswasCVPR2013,Parkash_ECCV_2012}. In contrast, we focus on machines providing justifications to humans. This could potentially allow machines to teach new concepts to humans (machine teaching). Other recent work~\cite{Hendricks_ECCV_2016} looks at post-hoc explanations for classification decisions. Instead of explaining why a model thinks an image is a particular class,~\cite{Hendricks_ECCV_2016} describes why an image is of a class predicted by the classifier. Unlike this task, our justification task requires reasoning about explicit context from the distractor class. Further, we are not interested in providing rationalizations for classification decisions but in explaining the differences between confusing concepts to humans. We show a comparison to~\cite{Hendricks_ECCV_2016} in the appendix, demonstrating the importance of context for justification.
\\[2pt]
\noindent{\textbf{Beam Search with Modified Objectives:}} Beam search is an approximate, greedy technique for inference in sequential models. We perform beam search on a modified objective for our introspective speaker model to induce discrimination. This is similar in spirit to recent works on inducing diversity in beam search~\cite{Vijayakumar_Arxiv_2016}, and maximum mutual information inference for sequence-to-sequence models~\cite{Li2016ADO}.

\section{Approach}\label{sec:approach}
\newcommand{\target}{c_t{}}
\newcommand{\distractor}{c_d{}}
\newcommand{\concept}{c{}}
\newcommand{\image}{I{}}

\rama{We describe our approach for inducing context-aware language for: 1) \emph{\justify{}}, where the context is another class, and 2) \emph{\disccap}, where the context is a semantically similar image. For clarity, we first describe the formulation for \justify, and then discuss a modification for \disccap.}

In the justification task (\reffig{fig:teaser} top), we wish to produce a sentence $s$, comprised of a sequence of words $\{s_i\}$, based on a given image $\image$ of a target concept $\target$ in the context of a distractor concept $\distractor$. The produced justification should capture aspects of the image that discriminate between the target, and the distractor concepts. Note that images of the distractor class are not provided to the algorithm.

We first train a generic context-agnostic image captioning model (from here on referred to as speaker) using training data from Reed~\etal~\cite{Reed_CVPR_2016} who collected captions describing bird images on the CUB-200-2011~\cite{WahCUB_200_2011} dataset. We condition the model on $\target$ in addition to the image. That is, we model $p(s|\image,\target)$. This not only helps produce better sentences (providing the model access to more information), but is also the cornerstone of our approach for discrimination~(\refsec{subsec:intro_speaker}). Our language models 
are recurrent neural networks which represent the state-of-the-art for language modeling across a range of popular tasks like image captioning~\cite{Vinyals_CVPR_2015,Xu_ICML_2015}, machine translation~\cite{Bahdanau_NIPS_2014} \etc.

\subsection{Reasoning Speaker}\label{subsec:reasoning_speaker}
To induce discrimination in the utterances from a language model, it is natural to consider using a generator, or speaker, which models $p(s|\image, \target$) in conjunction with a listener function  $f(s, \target, \distractor)$ 
that scores how discriminative an utterance $s$ is. The task of a pragmatic reasoning speaker $RS$, then, is to select utterances which are good sentences as per the generative model $p$, and are discriminative per $f$:
\begin{equation}\label{eqn:reasoning_speaker}
RS(\image, \target, \distractor)\!=\!\argmax_{s}
\lambda p(s| \image, \target) + 
(1\!-\!\lambda) f(s, \target, \distractor)
\end{equation}
where $0\leq\!\lambda\!\leq 1$ controls the tradeoff between linguistic adequacy of the sentence, and discriminativeness.

A similar reasoning speaker model forms the core of the approach of~\cite{pragma}, where $p$, and $f$ are implemented using multi-layer perceptrons (MLPs). As noted in~\cite{pragma}, selecting utterances from such a reasoning speaker poses several challenges. First, exact inference in this model over the exponentially large space of sentences is intractable. Second, in general one would not expect the discriminator function $f$ to factorize across words, making
joint optimization of the reasoning speaker objective difficult. Thus, Andreas, and Klein~\cite{pragma} adopt a sampling based strategy, where $p$ is considered as the proposal distribution whose samples are ranked by a linear combination of $p$, and $f$ (\refeq{eqn:reasoning_speaker}). Importantly, this distribution is over full sentences, hence the effectiveness of this formulation depends heavily on the distribution captured by $p$, since the
search over the space of all strings is solely based on the speaker. This is inefficient, especially when there is a mismatch in the statistics of the context-free (generative), and the unknown context-aware (discriminative) sentence distributions. \rama{In such cases, one must resort to drawing many samples to find good discriminative sentences.}

\subsection{Introspective Speaker}\label{subsec:intro_speaker}
Our approach for incorporating contextual behavior is based on a simple modification to the listener \emph{f} (\refeq{eqn:reasoning_speaker}). Given the generator $p$, we construct a listener module that wants to discriminate between $\target$, and $\distractor$, using the following log-likelihood ratio:
\vspace{-5pt}
\begin{equation}\label{eqn:introspector}
f(s, \target, \distractor)
= \log \frac{p(s| \target, \image)}{p(s| \distractor, \image)}.
\end{equation}
\vspace{-10pt}

This listener only depends on a generative model,
$p(s|c,\image)$, for the two classes $c_t$, and $c_d$.
We name it ``introspector'' to emphasize that this step re-uses the generative model, and does not need to train an explicit listener model.
Substituting the introspector into ~\refeq{eqn:reasoning_speaker} induces the following introspective speaker model for discrimination:
\vspace{-8pt}
\begin{multline}
\label{eqn:intro_speaker}
\underbrace{\Delta(\image, \target, \distractor)}_{\text{introspective speaker}} = \argmax_{s} \lambda \underbrace{\log p(s|\target, \image)}_{\text{speaker}}\\ \!\!\!\! + (1\! - \!\lambda) \underbrace{\log \frac{p(s|\target, \image)}{p(s| \distractor, \image)}}_{\text{introspector}}\quad,
\end{multline}
\vspace{-10pt}

with $\lambda$ that trades-off the weight given to generation, and introspection (similar to~\refeq{eqn:reasoning_speaker}). In general, we expect this approach to provide sensible results when $\target$, and $\distractor$ are similar. That is, we expect humans to describe similar concepts in similar ways, hence $p(s|\target, \image)$ should not be too different from $p(s|\distractor,\image)$. Thus, the introspector is less likely to overpower the speaker in \refeq{eqn:intro_speaker} in such cases (for a given $\lambda$). Note that for sufficiently different concepts the speaker alone is likely to be sufficient for discrimination. That is, describing the concept in isolation is likely to be enough to discriminate against a different or unrelated concept.

A careful inspection of the introspective speaker model reveals two desirable properties over previous work~\cite{pragma}. First, the introspector model does not need training, since it only depends on $p$, the original generative model. Thus, existing language models can be readily re-used to produce context-aware outputs by conditioning on $\distractor$. We demonstrate empirical validation of this in~\refsec{section:experiments}. This would help scale this approach to scenarios where it is not known apriori which concepts need to be discriminated, in contrast to approaches which train a separate listener module. \rama{Second, it leads to a unified, and efficient inference for the introspective speaker (\refeq{eqn:intro_speaker}), which we describe next.}

\subsection{Emitter-Suppressor (ES) Beam Search for RNNs}\label{subsec:emitter_suppressor_bs}

We now describe a search algorithm for implementing the maximization in~\refeq{eqn:intro_speaker}, which we call {\em{emitter-suppressor}} (ES) beam search. We use the beam search~\cite{beam_search} algorithm, which is a heuristic graph-search algorithm commonly used for inference in Recurrent Neural Networks~\cite{Hochreiter1997LongSM,Vijayakumar_Arxiv_2016}.

We first factorize the posterior log-probability terms in the introspective speaker equation (\refeq{eqn:intro_speaker}) $p(s|\target,\image) = \prod_{\tau=1}^{T} p(s_\tau|s_{1:\tau-1},\target,\image)$, denoting $s_{1:T}=\{s_\tau\}_{\tau=1}^{T}$ ($s_{1:0}$ corresponds to a null string). $T$ is the length of the sentence. We then combine terms from \refeq{eqn:intro_speaker}, yielding the following emitter-suppressor objective for the introspective speaker:
\begin{equation}\label{eqn:emitter_supressor}
\!\!\!\Delta(\image, \target, \distractor) = \argmax_{s} \sum_{\tau=1}^{T} \log \frac{\overbrace{p(s_\tau| s_{1: \tau-1}, \target, \image)}^{\text{emitter}}}{\underbrace{p(s_\tau| s_{1: \tau-1},\distractor, \image)^{1 - \lambda}}_{\text{suppressor}}}.
\end{equation}
The emitter (numerator in~\refeq{eqn:emitter_supressor}) is the generative model conditioned on the target concept $\target$, deciding which token to select at a given timestep. The suppressor (the denominator in~\refeq{eqn:emitter_supressor}) is conditioned on the distractor concept $\distractor$, providing signals to the emitter on which tokens to avoid. \rama{This is intuitive -- to be discriminative, we want to emit words that match $\target$, but avoid emitting words that match $\distractor$.}

\begin{figure}
	\vspace{-15pt}
\includegraphics[width=\columnwidth,page=2]{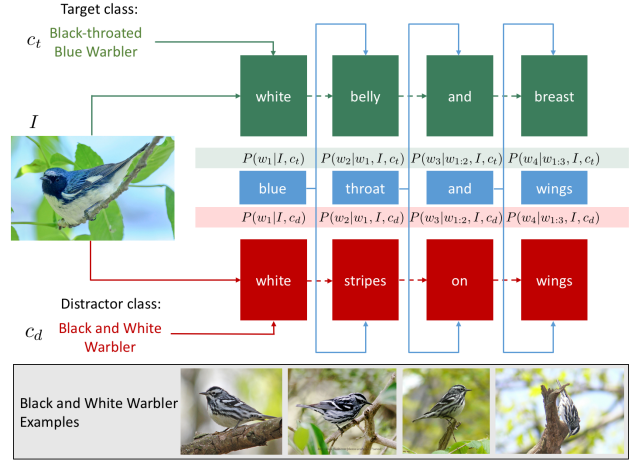}
\caption{
\footnotesize
Emitter-suppressor beam search for beam size 1, for distinguishing an image of ``black-throated blue warbler'' from the distractor class ``black and white warbler". Green: A language model $p(s|c_t, I)$ produces a caption ``white belly and breast ... ". Red: When feeding the distractor class to the language model, since the two birds share the attribute white belly, which appears in the image, the term "white" is highly suppressed. Blue: Picking likely words for the emitter, and unlikely for the suppressor yields a discriminative caption ``blue throat ..". Note that emitter, and suppressor share history (the previouly generated words).\vspace{-15pt}}
\label{fig:model_figure}
\end{figure}

We maximize the emitter-suppressor objective (\refeq{eqn:emitter_supressor}) using beam search. Vanilla beam search, as typically used in language models, prunes the output space at every time-step keeping the top-B (usually incomplete) sentences with highest log-probabilities so far (speaker in \refeq{eqn:intro_speaker}). Instead, we run beam search to keep the top-B sentences with highest ES ratio in~\refeq{eqn:emitter_supressor}. \reffig{fig:model_figure} illustrates this ES beam search for a beam size of 1.

It is important to consider how the trade-off parameter $\lambda$ affects the produced sentences. For $\lambda=1$, the model generates descriptions that ignore the context. At the other extreme, low $\lambda$ values are likely to make the produced sentences very different from any sentence in the training set (repeated words, ungrammatical sentences). 
It is not trivial to assume that there exists a wide enough range of $\lambda$ creating sentences that are both discriminative, and well-formed. However, our results (\refsec{sec:results}) indicate that such a range of $\lambda$ exists in practice.

\subsection{Discriminative Image Captioning}
\rama{We are given a target image $\image_t$, and a distractor $\image_d$},that we wish to \rama{distinguish}, similar to the two classes for the justification task. \rama{We construct a speaker (or generator) for this task by training a standard image captioning model.} Given this speaker, we construct an emitter-suppressor equation (as in~\refeq{eqn:emitter_supressor}):
\vspace{-5pt}
\begin{equation}\label{eqn:cap_emitter_supressor}
\!\!\!\Delta(\image_t, \image_d) = \argmax_{s} \sum_{\tau=1}^{T} \log \frac{\overbrace{p(s_\tau| s_{1: \tau-1}, \image_t)}^{\text{emitter}}}{\underbrace{p(s_\tau| s_{1: \tau-1},\image_d)^{1 - \lambda}}_{\text{suppressor}}}.
\end{equation}
\vspace{-10pt}

We re-use the mechanics of emitter-suppressor beam search from~\refsec{subsec:emitter_suppressor_bs}, conditioning the emitter on the target image $\image_t$, and the suppressor on the distractor image $\image_d$.


\section{Experimental Setup}\label{sec:experimental_setup}
\rama{We provide details of the CUB dataset, of our \cubjust{} dataset used for evaluation, and of the speaker-training setup for the \justify{} task. We then discuss the experimental protocols for \disccap{}.}

\subsection{Justification}\label{subsec:cub_justify}
\noindent{\textbf{CUB Dataset:}}
The Caltech UCSD birds (CUB) dataset~\cite{WahCUB_200_2011} contains 11788 images for 200 species of North American birds.
Each image in the dataset has been annotated with 5 fine-grained captions by Reed~\etal~\cite{Reed_CVPR_2016}. These captions mention various details about the bird (``This is a white spotted bird with a long pointed black beak.'') while not mentioning the name of the bird species.
\\[2pt]
\noindent\textbf{\cubjust{} Dataset:} \rama{We collect a new dataset (\cubjust{}) with ground truth justifications for evaluating \rama{justification}. We first sample the target, and distractor classes from within a hyper-category created based on the last name of the folk names of the 200 species in CUB. For instance, ``rufous hummingbird'', and ``ruby throated hummingbird'' both fall in the hyper-category ``hummingbird''. We induce 37 such hyper-categories. The largest single hypercategory is ``Warbler'' with 25 categories.
We then select a subset of (approx.) 15 images from the test set of CUB-200-2011~\cite{WahCUB_200_2011} for each of the 200 classes, to form a \cubjust{} test split. We use the rest for speaker training (\cubjust{} train split).}

\rama{Workers were then shown an image of the ``rufous hummingbird'', for instance, and a set of 6 other images (from \cubjust{} test split) all belonging to the distractor class ``ruby throated hummingbird'', to form the visual notion of the distractor class.}
They were also shown a diagram of the morphology of birds indicating various parts such as tarsus, rump, wingbars~\etc (similar to Reed~\etal~\cite{Reed_CVPR_2016}). The instruction was to describe the target image such that it is not confused with images from the distractor class. Some birds are best distinguished by non-visual cues such as their call, or their migration patterns.
Thus, we drop the categories of birds from the original list of triplets which were labeled as too hard to distinguish by the workers.
At the end of this process we are left with 3161 triplets with 5 captions each. We split this dataset into 1070 validation (for selecting the best value of $\lambda$), and 2091 test examples respectively. More details on the interface can be found in the appendix.
\\[2pt]
\noindent\textbf{Speaker Training:} We implement a model similar to ``Show, Attend, and Tell'' from Xu~\etal~\cite{Xu_ICML_2015}, modifying the original model to provide the class as input, similar in spirit to~\cite{Hendricks_ECCV_2016}. Exact details of our model architecture are given in the appendix. We train the model on the \cubjust{} train split. Recall that this just has context-agnostic captions from~\cite{Reed_CVPR_2016}.

To evaluate the quality of our speaker model, we report numbers here using the CIDEr-D metric~\cite{Vedantam_2015_CVPR} commonly used for image captioning~\cite{Hendricks_ECCV_2016,Karpathy_2015_CVPR,Vinyals_CVPR_2015} computed on the context-agnostic captions from~\cite{Reed_CVPR_2016}. Our captioning model with both the image, and class as input reaches a validation score of 50.2 CIDEr-D, while the original image-only captioning model reaches a CIDEr-D of 49.1. \rama{The scores are in a similar range as existing CUB captioning approaches~\cite{Hendricks_ECCV_2016}.}
\\[2pt]
\noindent\textbf{Justification Evaluation:}
We measure performance of the (context-aware) \justify{} captions on the \cubjust{} discriminative captions using the CIDEr-D metric. CIDEr-D weighs n-grams by their inverse document frequencies (IDF), giving higher weights to sentences having ``content" n-grams (``red beak'') than generic n-grams (``this bird'')~\cite{Hendricks_ECCV_2016}. Further, CIDEr-D captures importance of an n-gram for the image. For instance, it emphasizes ``red beak'' over, say, ``black belly'' if  ``red beak" is used more often in human justifications. We also report METEOR~\cite{meteor} scores for completeness. More detailed discussion on metrics can be found in the appendix.

\subsection{Discriminative Image Captioning}\label{subsec:setup/disc_image_caption}
\noindent{\textbf{Dataset:}} We want to test if reasoning about context with an introspective speaker can help discriminate between pairs of very similar images from the \coco{} dataset. To construct a set of confusing image pairs, we follow two strategies. First, \easy: For each image in the validation (test) set, we find its nearest neighbor in the FC7 space of a pre-trained VGG-16 CNN~\cite{DBLP:journals/corr/SimonyanZ14a}, and repeat this process of neighbor finding for 1000 randomly chosen source images. Second, \hard: To further narrow down to a list of semantically similar confusing images, we then run the speaker model on the nearest neighbor images, and compute word-level overlap (intersection over union) of their generated sentences. \rama{We then pick the top 1000 pairs with most overlap. Interestingly, the top 539 pairs had identical captions.} This reflects the issue of the output of image captioning models lacking diversity, and seeming templated~\cite{Delvin_Arxiv_2015,Vinyals_CVPR_2015}.
\\[2pt]
\noindent{\textbf{Speaker Training and Evaluation:}} We train our generative speaker for use in emitter-suppressor beam search using the model from~\cite{Vinyals_CVPR_2015} implemented in the neuraltalk2 project~\cite{kaprathy_neuratalk_2015}. We use the train/val/test splits from~\cite{Karpathy_2015_CVPR}. Our trained and finetuned speaker model achieves a performance of 91 CIDEr-D on the test set. As seen in~\refeq{eqn:cap_emitter_supressor}, no category information is used for this task. We evaluate approaches for discriminative image captioning based on how often they help humans to select the correct image out of the pair of images.

\section{Results}\label{sec:results}
\label{section:experiments}
\begin{figure}[htbp]
\includegraphics[width=\columnwidth]{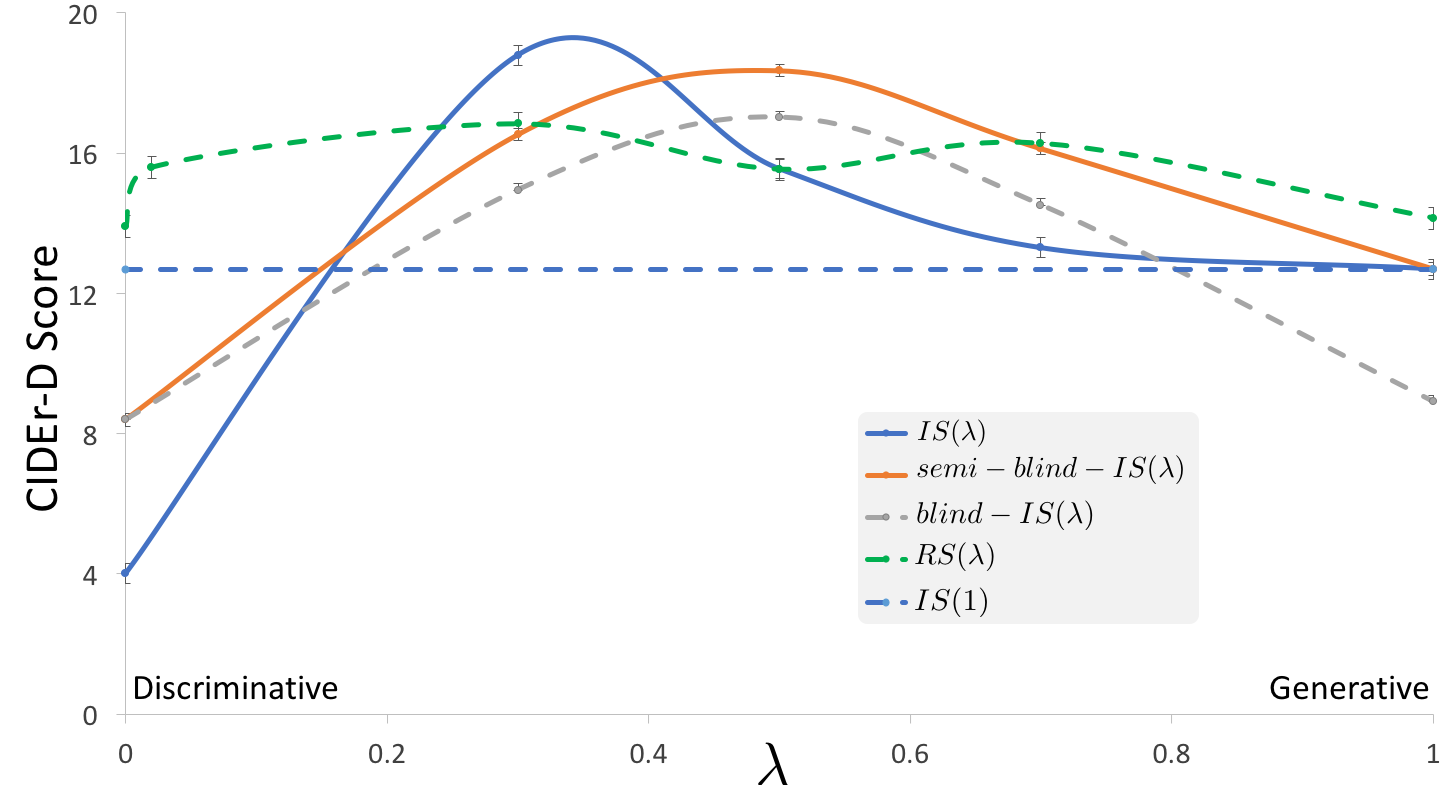}
\vspace{-15pt}
\caption{
{\bf{\cubjust{} validation results:}} CIDEr-D vs. $\lambda$ on \cubjust{} validation. Our introspective speaker approaches (\groundjust{} and \groundjustmixed) models perform best, followed by the class-only introspective speaker (\blindjust). \groundjustmixed{} outperforms other methods for a wider range of $\lambda$. All approaches which reason about pragmatics beat the baseline generative approach \literaljust. Error bars denote standard error of the mean score estimated across the validation set.
\vspace{-25pt}
}
\label{fig:inference_plot}
\end{figure}

\subsection{Justification}
\noindent{\textbf{Methods and Baselines:}} We evaluate the following models: 
{\bf 1.}~\groundjust: Introspective speaker from \refeq{eqn:intro_speaker};
{\bf 2.}~\literaljust: standard literal speaker, which generates a caption conditioned on the image and target class, but which ignores the distractor class;
 {\bf 3.}~\groundjustmixed: Introspective speaker in which the listener does not have access to the image, but the speaker does;
 {\bf 4.}~\blindjust: Introspective speaker without access to image, conditioned only on classes;
{\bf 5.}~\pragma: \rama{Our implementation of Andreas and Klein}~\cite{pragma}, but using 
our (more powerful) language model, and 
\refeq{eqn:intro_speaker} with a listener that models $\frac{p(s|c_t)}{p(s|c_d)}$ (similar to \groundjustmixed) for ranking samples (as opposed to a trained MLP~\cite{pragma}, to keep things comparable).
\rama{All approaches use 10 beams/samples (which is better than lower values) unless stated otherwise.}
\\[2pt]
\noindent{\textbf{Validation Performance:}}~\reffig{fig:inference_plot} shows the performance on \cubjust{} validation set as a function of $\lambda$, the hyperparameter controlling the tradeoff between the speaker and the introspector (\refeq{eqn:intro_speaker}). For the \pragma{} baseline, $\lambda$ stands for the tradeoff between the log-probability of the sentence and the score from the discriminator function for sample re-ranking. A few interesting observations emerge. First, both our \groundjust{} and \groundjustmixed{} models outperform the baselines for the mid range of $\lambda$ values. \groundjust{} model does better overall, but \groundjustmixed{} has a more stable performance over a wider range of $\lambda$. This indicates that when conditioned on the image, the introspector has to be highly discriminative (low lambda values) to overcome the signals from the image, since discrimination is between classes.

\begin{figure}[tbp]
	\includegraphics[width=\columnwidth]{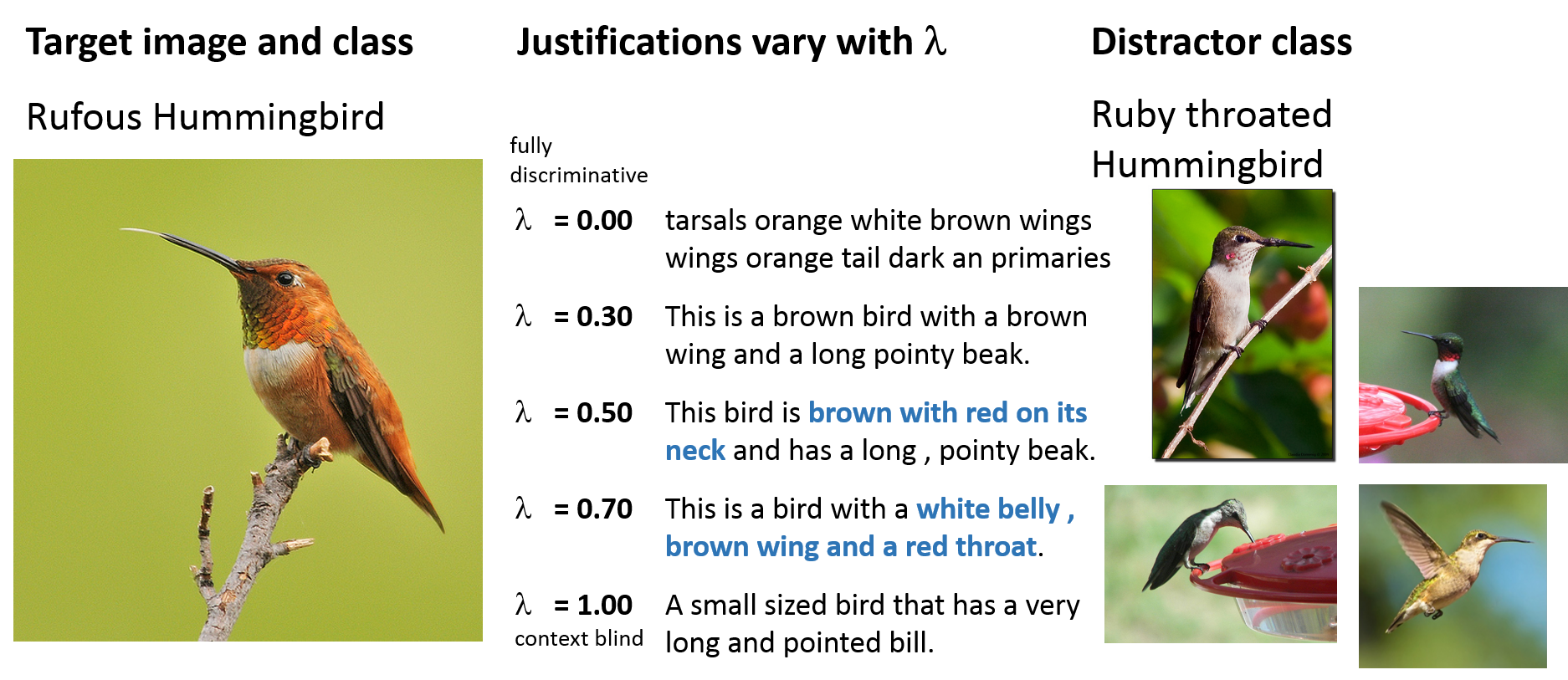}
	\caption{
		\footnotesize
		{\bf{The effect of context weight}}: An image of a ``Rufous Hummingbird'' in the context of another hummingbird type. A generative (context-blind) description describes the bird as having a long beak, but this feature is not discriminative. When taking into account the context, intermediate $\lambda$ values yield descriptions that highlight that the Rufous is brown with a red throat. For $\lambda=0$, the model does not force sentences to be well formed.}
	\label{fig:cub_across_lambda}
	\vspace{-10pt}
\end{figure}

Second, as $\lambda$ is decreased from 1, most methods improve as the sentences become more discriminative, but then get worse again as $\lambda$ becomes too low. This is likely to happen because when $\lambda$ is too low, the model explores rare tokens and parts of the output space that have not been seen during training, leading to badly-formed sentences~(\reffig{fig:cub_across_lambda}). \rama{This effect is stronger for~\groundjust{} models than for \pragma{}, since \pragma{} searches the output space over samples from the generator and only ranks using the joint reasoning speaker objective~(\refeq{eqn:reasoning_speaker}).}
Interestingly, at $\lambda=1$ (no discrimination), the \pragma{} approach, which samples from the generator, also performs better than other approaches, which use beam search to select high log-probability (context-agnostic) sentences. \rama{This indicates that in the absence of ground truth justifications, there is indeed a discrepancy between searching for discriminativeness and searching for a highly likely context-agnostic sentence.}

We perform more comparisons with the \pragma{} baseline, sweeping over $\{10, 50, 100\}$ samples from the generator for listener reranking (\refeq{eqn:reasoning_speaker}). We find that using 100 samples, \pragma{} gets comparable CIDEr-D scores (18.8) (but lower METEOR scores) than our \groundjustmixed{} approach with a beam size of 10. This suggests that our \groundjustmixed{} approach is more computationally efficient at exploring the output space because our emitter-suppressor beam search allows us to do joint greedy inference over speaker and introspector, leading to more meaningful local decisions. \rama{For completeness, we also trained a listener module discriminatively, and used it as a ranker for \pragma{}. We found that this gets to 16.2 $\pm$ 0.3 CIDEr-D (at $\lambda=0.5$) on validation, which is lower than \groundjust, showing that the bottleneck for performance is sampling, rather than the discriminativeness of the listener. More details can be found in the appendix.}
\\[2pt]
\noindent{\textbf{Test Performance:}} \reftab{table:inference_test} details the performance of the above models on the test set of \cubjust, with each model using its best-performing $\lambda$ on the validation set (\reffig{fig:inference_plot}). Both introspective-speaker models strongly outperform the baselines, with  \groundjustmixed{} slightly outperforming the \groundjust{} model. This could be due to the performance of \groundjustmixed{} being less sensitive to the exact choice of $\lambda$ (from~\reffig{fig:inference_plot}). Among the baselines, the best performing method is the \blindjust{} model, presumably because this model does emitter-suppressor beam search, \rama{while the other two baseline approaches rely on sampling and regular beam search respectively.}

\begin{table} 
\footnotesize
\setlength{\tabcolsep}{9pt}
\begin{center}
\begin{tabular}{@{} l  c  c  c @{}}
\toprule
Approach & CIDEr-D & \!\!METEOR\\
\midrule
\groundjust & 18.4 $\pm$ 0.2 & 26.5\\
\groundjustmixed & \textbf{18.5 $\pm$ 0.2} & \textbf{27.5} \\
\pragma & 15.8 $\pm$ 0.2 & 26.5 \\
\literaljust & 12.3 $\pm$ 0.1 & 25.3 \\
\blindjust & 16.1 $\pm$ 0.2 & 26.8 \\
\bottomrule
\end{tabular}
\caption{
\footnotesize
\textbf{\cubjust test results:} CIDEr-D, and METEOR scores (higher the better) computed on test set of \cubjust. Each model used the best $\lambda$ selected on the validation set (\reffig{fig:inference_plot}). Error values are standard error of the mean (SEM is less than 0.05 for METEOR). \groundjustmixed{} outperforms other methods.\vspace{-10pt}}
\label{table:inference_test}
\vspace{-25pt}
\end{center}
\end{table}

\begin{figure}[tbp]
\includegraphics[width=\columnwidth]{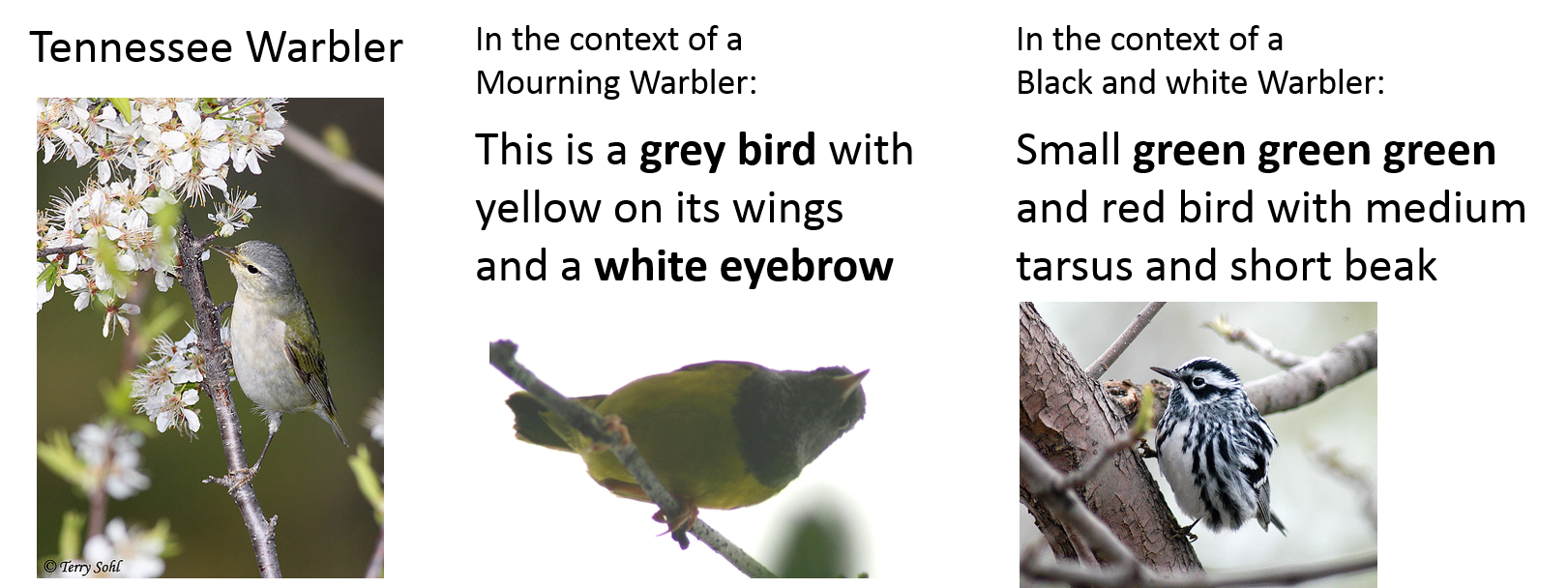}
	\caption{
    \footnotesize
    {\bf{The effect of context class}}: An image of a ``Tennessee Warbler'', which has light green wings, and a white eyebrow. When described in the context of a mourning warbler, which has a green hue, the description highlights that the target bird has a white eyebrow. When described in the context of the ``Black and White Warbler'', the description highlights that the target bird has green color.
    \vspace{-20pt}}
	\label{fig:cub_cross_category}
\end{figure}

\noindent{\textbf{Qualitative Results:}} We next showcase some qualitative results that demonstrate 1) aspects of pragmatics, and 2) context dependence captured by our best-performing \groundjustmixed{} model. \reffig{fig:cub_across_lambda} demonstrates how sentences uttered by the introspective speaker change with $\lambda$. At $\lambda=1$ the sentence describes the image well, but is oblivious of the context (distractor class). The sentence ``A small sized bird has a very long and pointed bill.'' is discriminative of hummingbirds against other birds, but not among hummingbirds (many of which tend to have long beaks/bills). At $\lambda=0.7$, and $\lambda=0.5$, the model captures discriminative features such as the ``red neck'', ``white belly'', and ``red throat''. Interestingly, at $\lambda=0.7$ the model avoids saying ``long beak'', a feature shared by both birds. \rama{Next, \reffig{fig:cub_cross_category} demonstrates how the selected utterances change based on the context.} \rama{A limitation of our approach is that, since the model never sees discriminative training data, in some cases it produces repeated words (``green green green'') when encouraged to be discriminative at inference time.}

\begin{figure}[tbp]
\includegraphics[width=\columnwidth]{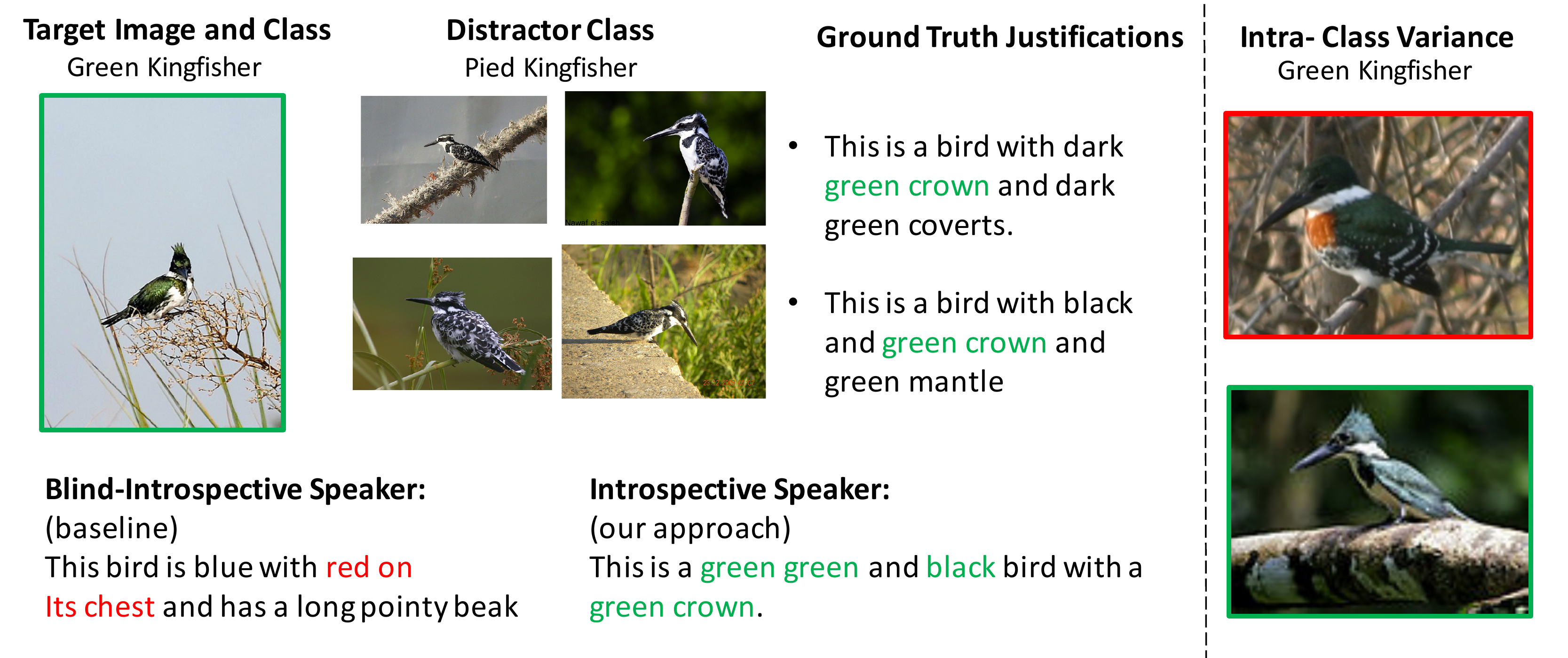}
\caption{
\footnotesize \textbf{The importance of visual signal for justification in fine-grained categories}. Given the image of a green kingfisher (left), a \blindjust{} model says the bird has ``red on its chest'',  which is inaccurate for this image, and a ``long pointy beak'', which is not a discriminative feature for this context. At the same time, the \groundjustmixed{} model mentions the ``green crown'', and avoids uttering ``red chest''. Given the complicated intra-category invariances in bird categories (right), it is intuitive that the image signal is important for justification.\vspace{-5pt}}
\label{fig:why_class}
\vspace{-5pt}
\end{figure}

\rama{Finally, \reffig{fig:why_class} illustrates the importance of visual reasoning for the justification task.} Fine-grained species often have large intra-class variances which a \emph{blind} approach to justification would ignore. Thus, a good justification approach needs to be grounded in the image signal to pick the discriminative cues appropriate for the given instance.

\begin{figure*}[tbp]
	\vspace{-5pt}
	\includegraphics[width=\textwidth]{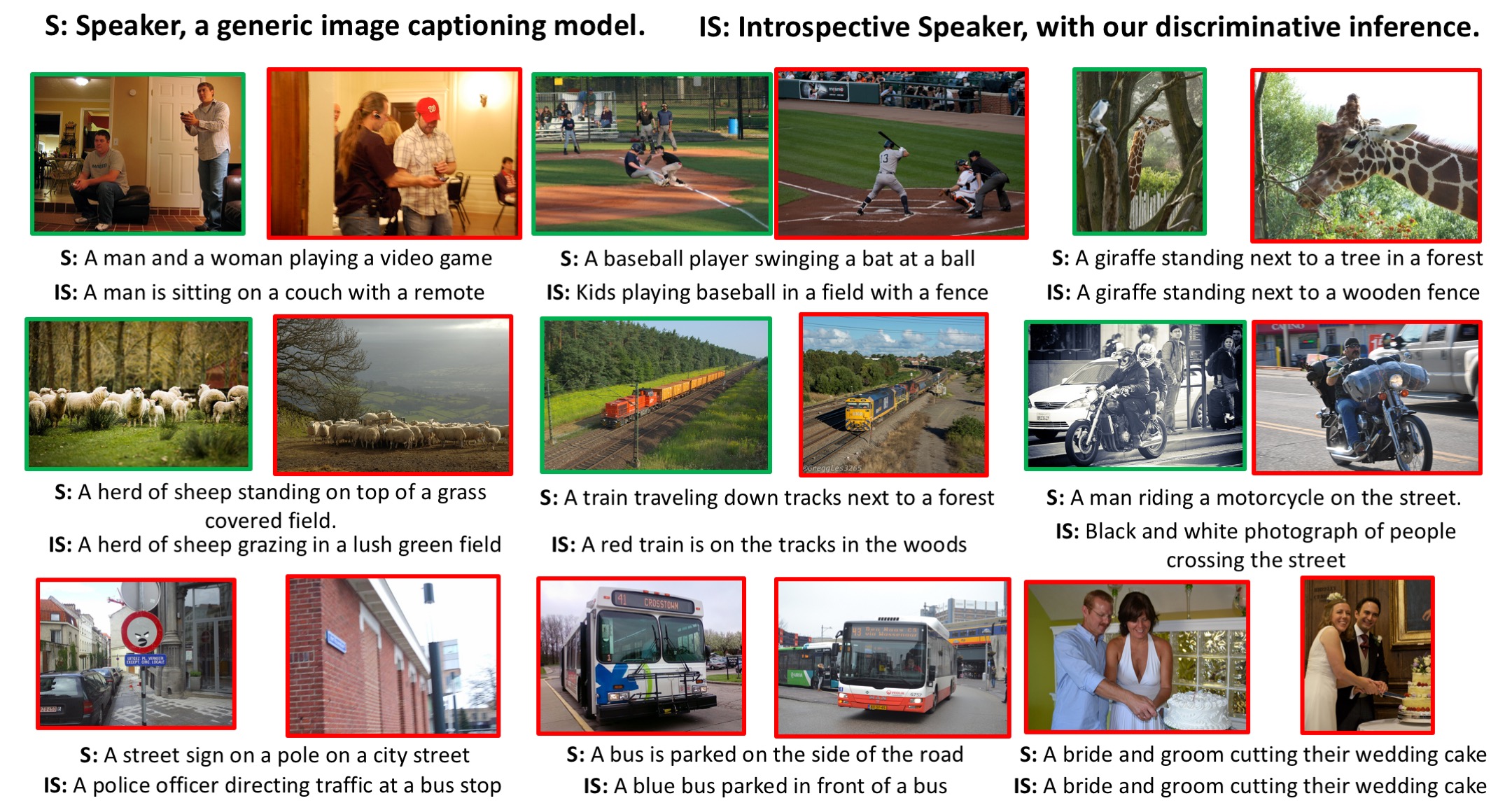}
	\caption{
		\footnotesize
		Pairs of images whose captions generated by a generic captioning speaker baseline (S) are identical. We apply our introspective speaker (IS) technique to distinguish the image on the left from the image on the right in each pair. The target image (left) is shown with a green border when the IS generated sentence is able to identify it correctly. Notice how the introspective speaker often refers more unambiguously to the target image. For example, for the sheep image (middle left), the IS generated sentence mentions that the sheep are grazing in a lush green field. In the bottom row we show some failure examples. The bottom left example is interesting, where the model calls the stop sign a policeman. In some cases (the wedding cake image), where the distributions captured by the emitter, and supressor RNN's are identical, our IS approach produces the same sentence as the baseline (S).
		\vspace{-15pt}
	}
	\label{fig:coco_qualitative}
\end{figure*}

\subsection{Discriminative Image Captioning}
As explained in \refsec{subsec:setup/disc_image_caption} we create two sets of semantically similar target, and distractor images: \easy{} based on FC7 features alone, and \hard{} based on both FC7, and sentences generated from the speaker (image captioning model). We are interested in understanding if emitter-suppressor inference helps identify the target image better than the generative speaker baseline. Thus the two approaches are speaker (S) (baseline), and introspective speaker (IS) (our approach). We use $\lambda=0.3$ based on our results on the CUB dataset. 
We run all approaches at a beam size of 2 (typically best for \coco~\cite{kaprathy_neuratalk_2015}).
\\[2pt]
\noindent{\textbf{Human Studies:}} We setup a two annotation forced choice (2AFC) study where we show a caption to raters asking them to ``pick an image that the sentence is more likely to be describing.''. Each target distractor image pair is tested against the generated captions. We check the fraction of times a method caused the target image to be picked by a human. A discriminative image captioning method is considered better if it enables humans to identify the target image more often. Results of the study are summarized in~\reftab{table:coco_human_studies}. We find that our approach outperforms the baseline speaker (S) on the \emph{easy confusion} as well as the \emph{hard confusion} splits. However, the gains from our approach are larger on the \emph{hard confusion} split, which is intuitive.
\\[2pt]
\noindent{\textbf{Qualitative Results:}} The qualitative results from our \coco{} experiments are shown in \reffig{fig:coco_qualitative}. The target image, when successfully identified, is shown with a green border. We show examples where our model identifies the target image better in the first two rows, and some failure cases in the third row. Notice how the model is able to modify its utterances to account for context, and pragmatics, when going from $\lambda=1$ (speaker) to $\lambda=0.3$ (introspective speaker). Note that the sentences typically respect grammatical constructs despite being forced to be discriminative.

\begin{table} \footnotesize
\setlength{\tabcolsep}{1.5pt}
\begin{center}
\begin{tabular}{@{} l  c  c  c @{}}
\toprule
Approach & \emph{easy confusion} (\%) & \emph{hard confusion} (\%)\\
\midrule
S (baseline) & 74.6 &  52.5\\
IS (ours) & \textbf{89.0} & \textbf{74.1} \\
\bottomrule
\end{tabular}
\caption{
\footnotesize \% of image pairs that are correctly discriminated by humans, based on descriptions in \coco. Introspective speaker (IS) is better at pointing to the target image given a confusing distractor image across both easy, and hard data splits than a speaker (S). Standard error is below the precision we report numbers at.\vspace{-10pt}}
\label{table:coco_human_studies}
\vspace{-20pt}
\end{center}
\end{table}

\section{Discussion}\label{sec:discussion}

Describing absence of concepts and inducing comparative language are exciting directions for future work on justification. For instance, when justifying why an image is a lion and not a tiger, it would be useful to be able to say ``because it does not have stripes.'', or ``because it has a more hair on its face.'' Beyond pragmatics, the justification task also has interesting relations to human learning. Indeed, we all experience that we learn better when someone takes time out to justify or explain their point of view. One can imagine such justifications being helpful for ``machine teaching'', where a teacher (machine) can provide justifications to a human learner explaining the rationale for an image belonging to a particular fine-grained category as opposed to a different, possibly mistaken, or confusing fine-grained category.

There are some fundamental limitations to inducing context-aware captions from context-agnostic supervision. For instance, if two distinct concepts are very similar, human-generated context-free descriptions may be identical, and our model (as well as baselines) would fail to extract any discriminative signal. Indeed, it is hard to address such situations without context-aware ground truth.

We believe modeling higher-order reasoning (such as pragmatics) by reusing the sampling distribution from language models can be a powerful tool. It may be applicable to other higher-order reasoning, without necessarily setting up policy gradient estimators on reward functions. Indeed, our inference objective can also be formulated for training. However, initial experiments on this did not yeild significant performance improvements.

\section{Conclusion}
We introduce a novel technique for deriving pragmatic language from recurrent neural network language models, namely, an image-captioning model that takes into account the context of a distractor class or a distractor image. Our technique can be used at inference time to better discriminate between concepts, without having seen discriminative training data. We study two tasks in the vision, and language domain which require pragmatic reasoning: \emph{justification} -- explaining why an image belongs to one category as opposed to another, and \emph{discriminative image captioning} -- describing an image so that one can distinguish it from a closely related image. Our experiments demonstrate the strength of our method over generative baselines, as well as adaptations of previous work to our setting. We will make the code, and datasets available online.

\footnotesize
\noindent\textbf{Acknowledgements:}
We thank Tom Duerig for his support, and guidance in shaping this project. We thank David Rolnick, Bharadwaja Ghali, Vahid Kazemi for help with~\cubjust{} dataset. We thank Ashwin Kalyan for sharing a trained checkpoint for the discriminative image captioning experiments. We also thank Stefan Lee, Andreas Veit, Chris Shallue. This work was funded in part by an NSF CAREER, ONR Grant N00014-16-1-2713, ONR YIP, Sloan Fellowship, ARO YIP, Allen Distinguished Investigator, Google Faculty Research Award, Amazon Academic Research Award to DP.

\normalsize

\section*{Appendix}

\setcounter{section}{0}
We organize the appendix as follows:
\begin{packed_itemize}
\item \refsec{sec:qualitative}: Analysis of performance as we consider unrelated images as distractors.
\item \refsec{sec:gve}: Generating visual explanations~\cite{Hendricks_ECCV_2016} adapted to the justification task.
\item \refsec{sec:show_attend_and_tell}: Architectural changes to the ``Show, Attend, and Tell'' image captioning model~\cite{Xu_ICML_2015} for justification.
\item \refsec{sec:optimization}: Optimization details for justification speaker model.
\item \refsec{sec:metrics}: Choice of metrics for evaluating justification.
\item \refsec{sec:dataset}: \cubjust{} data collection details.
\item \refsec{sec:misc}: Analysis of the~\pragma{} baseline in more detail.
\item \refsec{sec:trained_listener}: Comparison of our approach to a baseline with a discriminatively trained listener used for reranking in \pragma{} model.
\end{packed_itemize}

\section{COCO Qualitative Results}\label{sec:qualitative}
\begin{figure*}[htbp]
\centering
\includegraphics[width=0.84\textwidth]{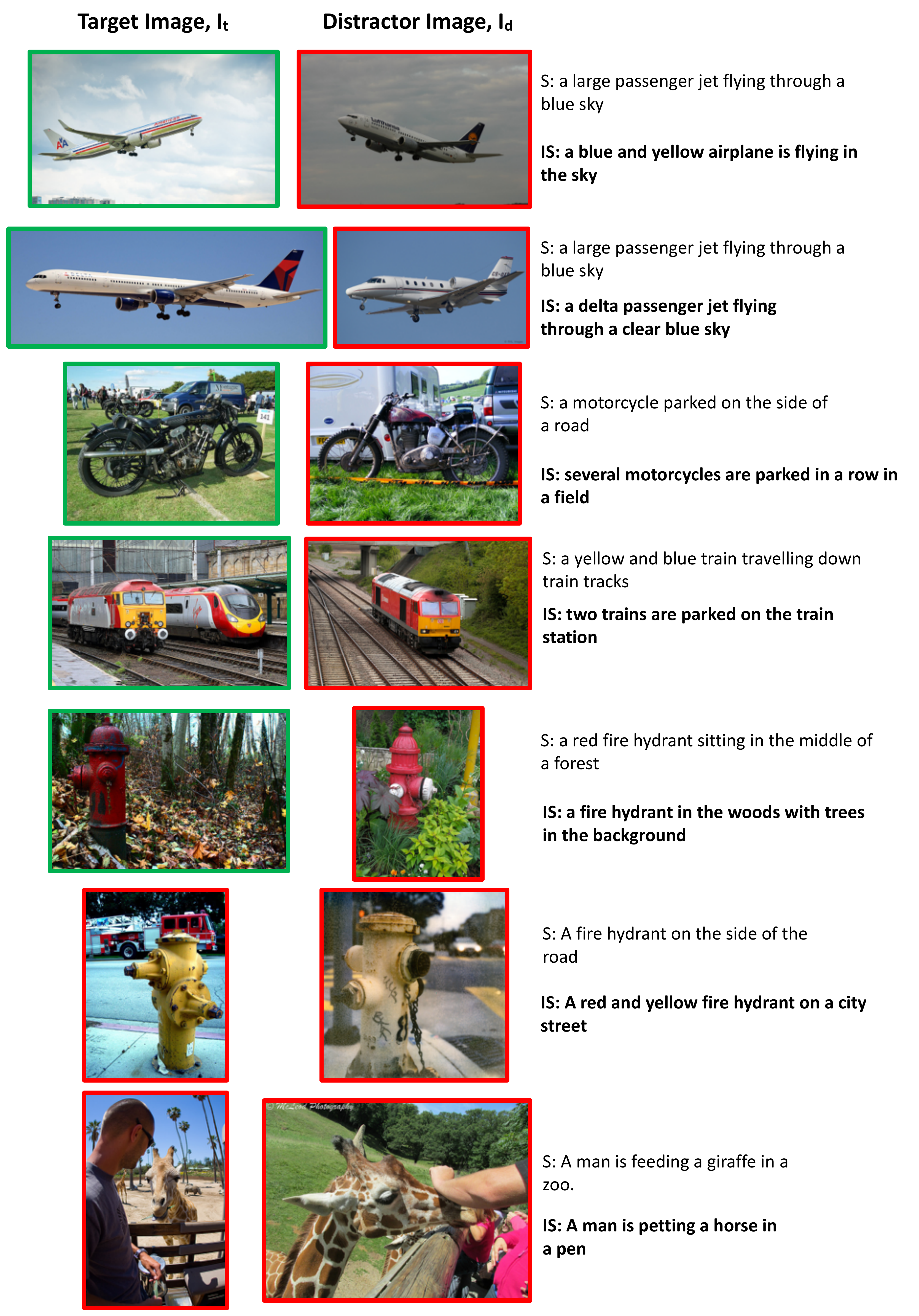}
\caption{Qualitative examples for discriminative image captioning (similar to~\reffig{fig:coco_qualitative}). S (speaker) denotes examples from the standard image captioning model, which generates the same caption for the two images. Our method's outputs are shown as IS (introspective speaker). The target image is shown to the left and marked with a green border where our approach is accurate, as well as more discriminative. The second last example shows a case where our model is more discriminative, but inaccurate for the original target image and the last example shows a case where our caption is neither accurate not discriminative.}
\label{fig:coco_qualitative_appendix}
\end{figure*}

\noindent{\textbf{COCO Qualitative Examples:}}~\reffig{fig:coco_qualitative_appendix} shows more qualitative results on discriminative image captioning on the \textbf{hard confusion} split of the \coco{} dataset. Notice how our introspective speaker captions (denoted by IS), which model the context (distractor image) explicitly are often more discriminative, helping identify the target image more clearly than the baseline speaker approach (denoted by S). For example in the second row, our IS model generates the caption ``a delta passenger jet flying through a clear blue sky", which is a more discriminative (and accurate) caption than the baseline caption ``a large passenger jet flying through a blue sky'', which applies to both the target and distractor images.

\begin{figure*}[htbp]
\includegraphics[width=\textwidth]{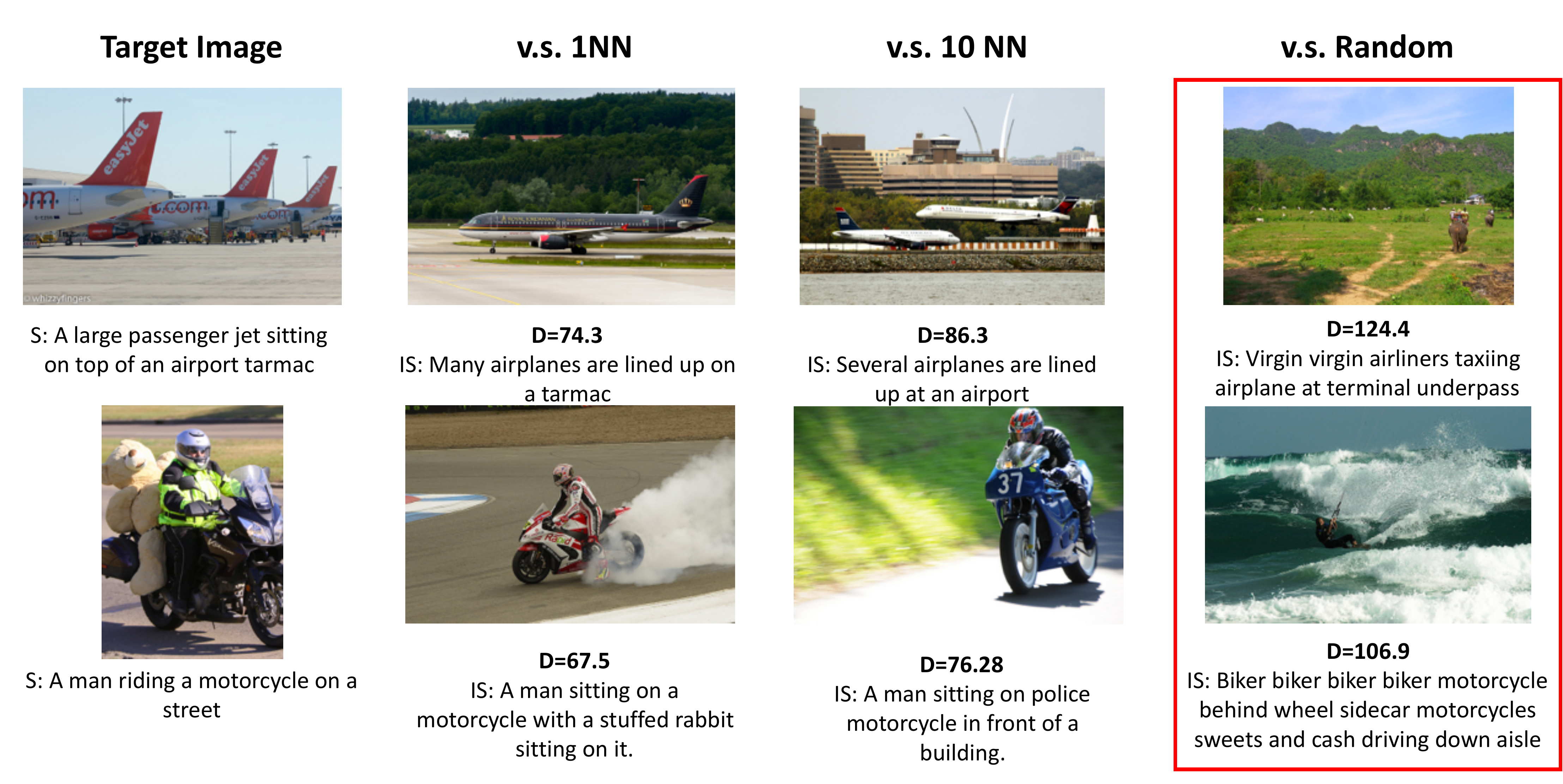}
\caption{We show the target image (extreme left) and distractor images at varying distances (1 nearest neighbor, 10 nearest neighbor and random distractor), along with some generated captions. D denotes the distance between the target and distractor images in the FC7 space.
	The output of the speaker (S) is shown under the target image and the output of the introspective speaker considering each distractor image as context in turn, is shown under the corresponding distractor image. 
That is, the caption under each distractor image describes the target image distinguishing it from the distractor. 
Notice that our introspective speaker (IS) method often works well for 1 nearest neighbour and the 10$^{th}$ nearest neighbor, but produces incomprehensible sentences when the distractor is irrelevant. Indeed, for a random distractor, we see that the baseline speaker outputs (S) are often sufficient for discrimination, which is intuitive.
}
\label{fig:random_coco_distractor}
\end{figure*}

\noindent{\textbf{Effect of increasing distance:}} We illustrate how the quality of the discriminative captions from the introspective speaker (IS) approach varies as the distractor image becomes less relevant to the target image (\reffig{fig:random_coco_distractor}). For the target image on the left, we show the 1-nearest neighbor (which has a very similar caption to the target image), the 10$^{th}$-nearest neighbor and a randomly selected distractor image. When we pick a random image to be the distractor, the generated discriminatve captions become less comprehensible, losing relevance as well as grammatical structure. This is consistent with our understanding of the introspective speaker (IS) formulation from~\refsec{subsec:intro_speaker}: modeling the context explicitly during inference helps discrimination when the context is relevant. When the context is not relevant, as with the randomly picked images, the original speaker model (S) is likely sufficient for discrimination. 

\section{Comparison to previous work on Generating Visual Explanations~\cite{Hendricks_ECCV_2016}}\label{sec:gve}
Hendricks~\etal~\cite{Hendricks_ECCV_2016} propose a method to explain classification decisions to an end user by providing post-hoc rationalizations. Given a prediction from a classifier, this work generates a caption conditioned on the predicted class, and the original image. While Hendricks~\etal aim to provide a rationale for a classification, we focus on a related but different problem of concept justification. Namely, we want to explain why an image contains a target class as opposed to a specific distractor class, while Hendircks~\etal want to explain why a classifier thought an image contains a particular class. Thus, unlike the visual explanation task, it is intuitive that the justification task requires explicit reasoning about context. We verify this hypothesis, by first adapting the work of~\cite{Hendricks_ECCV_2016} to our justification task, using it as a speaker, and then augmenting the speaker with our approach to construct an intropsective speakerm which accounts for context. Interestingly, we find that our introspective speaker approach helps improve the performance of generating visual explanations~\cite{Hendricks_ECCV_2016} on justification.

The approach of Hendricks~\etal~\cite{Hendricks_ECCV_2016} differs from our setup in two important ways. Firstly, uses a stronger CNN, namely the fine-grained compact-bilinear pooling CNN~\cite{Gao_CVPR_2016} which provides state-of-the-art performance on the CUB dataset. Secondly, to make the explanations more grounded in the class information, they also add a constraint to induce captions which are more specific to the class. This is achieved by using a policy gradient on a reward function that models $p(c| s)$ for a given sentence $s$ and class $c$. Thus, in some sense the approach encourages the model to produce sentences that are highly discriminative of a given class against all other classes, as opposed to a particular distractor class that we are interested in for justification. Finally, the policy gradient is used in conjunction with standard maximum likelihood training to train the explanation model. At inference, the explanation model is run by conditioning the caption generation on the predicted class.

We modify the inference setup of~\cite{Hendricks_ECCV_2016} slightly to condition the caption generation on the~\emph{target} class for justification, as opposed to the predicted class for explanation. We call this the \textbf{vis-exp} approach. We then apply the emitter-suppressor beam search (at a beam size of 1, to be consistent with~\cite{Hendricks_ECCV_2016}) to account for context, giving us an introspective visual explanation model (\textbf{vis-exp-IS}). Given the stronger image features and a more complicated training procedure involving policy gradients (hard to implement and tune in practice), the \textbf{vis-exp} approach achieves a strong CIDEr-D score of $20.36$ with a standard error of $0.16$ on our \cubjust{} test set. Note that this \cubjust{} test set is a strict subset of the test set from~\cite{Hendricks_ECCV_2016}. These results are better than those achieved with our~\groundjustmixed{} CUB model, which is based on regular image features from VGG-16 implemented in the ``Show, Attend and Tell'' framework and uses standard log-likelihood training (\reftab{table:inference_test}).

However, as mentioned before, the approach of~\cite{Hendricks_ECCV_2016}, similar to a baseline speaker S, cannot explicitly model context from a specific distractor class at inference. That is, while the approach reasons (through its training procedure) that given an image of a hummingbird, one should talk about its \emph{long beak} (a discriminating feature for a hummingbird against all other birds), it cannot reason about a specific distractor class presented at inference. If the distractor class is another hummingbird with a long beak, we would want to avoid talking about the \emph{long beak} in our justification. On the other hand, if the distractor class were a hummingbird with a shorter beak and there do exist such hummingbirds, then the \emph{long beak} would be an important feature to mention in a justification. Clearly, this is non-trivial to realize without explicitly modeling context. Hence, intuitively, one would expect that incorporating context from the distractor class should help the justification task.

\begin{table}
\footnotesize
\setlength{\tabcolsep}{9pt}
\begin{center}
\begin{tabular}{@{} l  c  c  c @{}}
\toprule
Approach & CIDEr-D\\
\midrule
\textbf{vis-exp}~\cite{Hendricks_ECCV_2016} & 20.36 $\pm$ 0.16\\
\textbf{vis-exp-IS} (ours) & 21.52 $\pm$ 0.17\\
\bottomrule
\end{tabular}
\caption{
\footnotesize
\textbf{\cubjust test results:} We compare \textbf{vis-exp}~\cite{Hendricks_ECCV_2016} and our emitter-suppressor beam search implemented on top of \textbf{vis-exp}, namely \textbf{vis-exp-IS}. We see that we can achieve gains over the \textbf{vis-exp} approach by explicitly reasoning about context using our introspective speaker on the justification task. Error values are standard error of the mean.}
\label{table:gve_table}
\end{center}
\vspace{-20pt}
\end{table}

As explained previously, we implement our emitter-suppressor inference (\refeq{eqn:intro_speaker}), on top of the \textbf{vis-exp} approach, yielding an \textbf{vis-exp-IS} approach. We sweep over the values of $\lambda$ on validation and find that the best performance is achieved at $\lambda=0.9$. Plugging this value and evaluating on test, our \textbf{vis-exp-IS} approach achieves a CIDEr-D score of $21.52$ with a standard error of $0.17$ (\reftab{table:gve_table}). This is an improvement of 1.16 CIDEr-D. Our gains over \textbf{vis-exp} are lower than the gains on the \literaljust{} approach (reported in~\reftab{table:inference_test}), presumably because the \textbf{vis-exp} approach already captures a lot of the context-independent discriminative signals (\eg, \emph{long beak} for a hummingbird), due to policy gradient training. Overall though, these results provide further evidence that our emitter-suppressor inference scheme can be adapted to a variety of context-agnostic captioning models, to effectively induce context awareness during inference.

\section{CUB Captioning Model Architecture}\label{sec:show_attend_and_tell}
We explain some minor modifications to the ``Show, Attend and Tell''~\cite{Xu_ICML_2015} image captioning model to condition it on the class label in addition to the image, for our experiments on CUB. Note that the explanation in this section is only for CUB -- our \coco{} models are trained using the neuraltalk2 package\footnote{https://github.com/karpathy/neuraltalk2} which implements the ``Show and Tell'' captioning model from Vinyals~\etal~\cite{Vinyals_CVPR_2015}. Our changes can be understood as three simple modifications aimed to use class information in the model. We first embed the class label (1 out of 200 classes for CUB) into a continuous vector $\mathbf{k} \in \mathbb{R}^D$, $D=512$. The three changes then, on top of the Show, Attend, and Tell model~\cite{Xu_ICML_2015} are as follows:
\begin{packed_itemize}
	\item \textbf{Changes to initial LSTM state:} The original Show, Attend, and Tell model uses image annotation vectors $a_i$ ($i$ indexes spatial location), which are the outputs from a convolutional feature map to compute the initial cell and hidden states of the long-short term memory (LSTM) ($c_0, h_0$). The image annotation vector is averaged across spatial locations $\mathbf{\bar{a}} = \frac{1}{L}\sum_{i=1}^L \mathbf{a_i}$ and used to compute the initial state as follows:
$$
\mathbf{c_0} = f_{init,c}(\mathbf{\bar{a}})
$$
$$
\mathbf{h_0} = f_{init,h}(\mathbf{\bar{a}})
$$
We modify this to also use the class embedding $k$ to predict the initial state of the LSTM, by concatenating it with the averaged anntoation vector ($\mathbf{\bar{a}}$):
$$
\mathbf{c_0} = f_{init,c}(\left[\mathbf{\bar{a}}; \mathbf{k}\right])
$$
$$
\mathbf{h_0} = f_{init,h}(\left[\mathbf{\bar{a}}; \mathbf{k}\right])
$$

\item \textbf{Changes to the LSTM recurrence:} ``Show, Attend and Tell'' computes a scalar attention $\alpha_{ti}$ at each location of the feature map and uses it to compute a context vector at every timestep $\mathbf{\hat{z_t}} = \phi(\{\alpha_{ti}, \mathbf{a_i}\})$ by attending on the image annotation $a_i$. It also embeds an input word $y_t$ using an embedding matrix $E$ and uses the previous hidden state $h_t$ to compute the following LSTM recurrence at every timestep, producing outputs $\mathbf{i_t}$ (input gate), $\mathbf{f_t}$ (forget gate), $\mathbf{o_t}$ (output gate), $\mathbf{g_t}$ (input) (Eqn. 1, 2, 3 from~\cite{Xu_ICML_2015}):
\begin{equation}
	\begin{split}
		\begin{pmatrix}
        \mathbf{i_t}\\
        \mathbf{f_t}\\
        \mathbf{o_t}\\
        \mathbf{g_t}\\
        \end{pmatrix}
        = \begin{pmatrix}
        \sigma\\
        \sigma\\
        \sigma\\
        \text{tanh}\\
        \end{pmatrix} T \begin{pmatrix}E \mathbf{y_t}\\ \mathbf{h_{t-1}}\\ \mathbf{\hat{z_t}}\end{pmatrix}
	\end{split}
\end{equation}
\begin{align}
\mathbf{c_t} &= \mathbf{f_t} \odot \mathbf{c_{t-1}} + \mathbf{i_t} \odot \mathbf{g_t}\\
\mathbf{h_t} &= \mathbf{o_t} \odot \text{tanh}(\mathbf{c_t})
\end{align}
We use the class embeddings $\mathbf{k}$ in addition to the context vector $\mathbf{\hat{z_t}}$ in Eqn. 1:
\begin{equation}
	\begin{split}
		\begin{pmatrix}
        \mathbf{i_t}\\
        \mathbf{f_t}\\
        \mathbf{o_t}\\
        \mathbf{g_t}\\
        \end{pmatrix}
        = \begin{pmatrix}
        \sigma\\
        \sigma\\
        \sigma\\
        \text{tanh}\\
        \end{pmatrix} T' \begin{pmatrix}E \mathbf{y_t}\\ \mathbf{h_{t-1}}\\ \mathbf{\hat{z_t}}\\ \mathbf{k}\end{pmatrix}
	\end{split}
\end{equation}
The remaining equations for the LSTM recurrence remain the same (Eqn. 2, 3). 
\item \textbf{Adding class information to the deep output layer:} ``Show, Attend and Tell'' uses a deep output layer~\cite{Pascanu_ICLR_2014} to compute the output word distribution at every timestep, incorporating signals from the LSTM hidden state $h_t$, context vector $\mathbf{\hat{z_t}}$ and the input word $\mathbf{y_t}$:
$$
p(y_t) \propto \exp (L_o(E\mathbf{y_t} + L_h \mathbf{h_t} + L_z \mathbf{z_t}))
$$
Here $L_h$, $L_z$ are matrices used to project $\mathbf{h_t}$ and $\mathbf{z_t}$ to the dimensions of the word embeddings $E \mathbf{y_t}$ and $L_o$ is the output layer which produces an output of the size of the vocabulary. Similar to the previous two adaptations, we use the class embedding $\mathbf{k}$ in addition to the context vector $\mathbf{\hat{z_t}}$ to predict the output at every timestep:
$$
p(y_t) \propto \exp (L_o(E\mathbf{y_t} + L_h \mathbf{h_t} + L_z \mathbf{z_t} + L_k \mathbf{k}))
$$
\item \textbf{Blind models:} For implementing our class-only \blindjust{} model, we need to train a model that only uses the class to produce a sentence. For this, we drop the attention component from the model, which is equivalent to setting $\hat{z_t}$ and $\hat{\bar{a}}$ to zero for all our equations above and run the model using the class embedding $\mathbf{k}$.
\end{packed_itemize}

\section{Optimization Details}\label{sec:optimization}
Our CUB captioning network is trained using \emph{Rmsprop}~\cite{Rmsprop} with a batch size of $32$ and a learning rate of $0.001$. We decayed the learning rate on every $5$ epochs of cycling through the training data. Our word embedding $E$ embeds words into a $512$ dimensional vector and we set LSTM hidden and cell state ($h_0, c_0$) sizes to 1800, similar to the ``Show, Attend, and Tell'' model on \coco{}. The rest of our design choices closely mirror the original work of~\cite{Xu_ICML_2015}, based on their implementation available at~\url{https://github.com/kelvinxu/arctic-captions}. We will make our Tensorflow implementation of ``Show, Attend, and Tell'' publicly available.

\section{Metrics for Justification}\label{sec:metrics}
In this section, we expand more on our discussion on the choice of metrics for evaluating justification (\refsec{subsec:cub_justify}). In addition to the metrics we report in the main paper, namely CIDEr-D~\cite{Vedantam_2015_CVPR} and METEOR~\cite{meteor}, we also considered using the recently introduced SPICE~\cite{Anderson_ECCV_2016}. The SPICE metric uses a dependency parser to extract a scene graph representation for the candidate and reference sentences and computes an F-measure between the scene graph representations. Given that the metric uses a dependency parser as an intermediate step, it is unclear how well it would scale to our justification task: some of the sentences from our model might be good justifications but may not be exactly grammatical. This is because our discriminative justifications emerge as a result of a tradeoff between high-likelihood sentences and discrimination (\refeq{eqn:intro_speaker}). Note that this tradeoff is inherent since we don't have ground truth (well-formed) discriminative training data. Thus SPICE can be a problematic metric to use in our context. However, for the sake of completeness, we report SPICE numbers on validation, giving each approach access to its best $\lambda$ value, in~\reftab{table:inference_val}.
\begin{table} 
\footnotesize
\setlength{\tabcolsep}{9pt}
\begin{center}
\begin{tabular}{@{} l  c  c  @{}}
\toprule
Approach & SPICE\\
\midrule
\groundjust & \textbf{16.45 $\pm$ 0.12}\\
\groundjustmixed & 15.59 $\pm$ 0.12\\
\pragma & 14.69 $\pm$ 0.12\\
\literaljust & 14.74 $\pm$ 0.12\\
\blindjust & 15.7 $\pm$ 0.12\\
\bottomrule
\end{tabular}
\caption{
\footnotesize
\textbf{\cubjust validation results:} SPICE scores (higher the better) computed on validation set of \cubjust. Each model used its best $\lambda$ value. Error values are standard error of the mean. \groundjust\ outperforms the other methods by a good margin on SPICE.}
\label{table:inference_val}
\vspace{-10pt}
\end{center}
\end{table}

Although we outperform the baselines using the SPICE metric, in some corner cases we also found the SPICE metric scores to be slightly un-interpretable. For example, for the candidate sentence ``this bird has a speckled belly and breast with a short pointy bill.'', and reference sentences ``This bird has a yellow eyebrow and grey auriculars'', ``This is a bird with  yellow supercilium and white throat'', the SPICE scores were higher than one would expect (0.30). For reference, an intuitively more related sentence ``this is a grey and yellow bird with a yellow eyebrow.'' obtains a lower SPICE score of 0.28 for the same reference sentences. Further investigation revealed that the relation F-measure, which roughly measures if the two sentences encode the same relations, had a high score in these corner cases. We hypothesize that this inconcsistency in scores might be because SPICE uses soft similarity from WordNet for computing the F-measure, which might not be calibrated for this fine-grained domain, with specialized words such as \emph{supercilium}, \emph{auriculars} \etc. As a result of these observations, we decided not to perform key evaluations with the SPICE metric.

\section{\cubjust{} Dataset Interface}\label{sec:dataset}
\begin{figure}[htbp]
\includegraphics[width=\columnwidth]{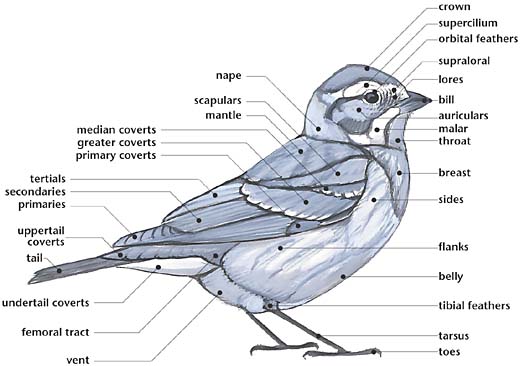}
\caption{A diagram of the morphology of a bird, labeling different parts. This diagram was shown to workers when getting justifications explaining why the image contains a target class, and not a distractor class.}
\label{fig:bird_morpohology}
\end{figure}

We provide more details on the collection of the \cubjust{} dataset (\refsec{subsec:cub_justify}). We presented a target image from a selected target class to the workers along with a set of six distractor images, all belonging to one other distractor class. The distractor images were chosen at random from the validation, and test split of the CUB dataset we created for justification. Non-expert workers are unlikely to given have an explicit visual model of a given ditractor category, say Indigo Bunting. Thus the distractor images were shown to entail the concept of the distractor class for justification. As explained in~\refsec{subsec:cub_justify} the choice of the distractor classes is made based on the hierarchy we induce using the folk names of the birds. Given the target class, and the distractor class images, workers were asked to describe the target image in a manner that the sentence is not confusing with respect to the distractor images. Further, the workers were instructed that someone who reads the sentence should be able to recognize the target image, distinguishing it from the set of distractor images. In order to get workers to pay attention to all the images (and the intra-class invariances), they were not told explicitly that the distractor images all belonged to one other, unique, distractor class. For helping identify minute difference between images of birds, as well as enabling workers to write more accurate captions, we also showed them a diagram of the morphology of a bird (\reffig{fig:bird_morpohology}). We also showed them a list of some other parts with examples not shown in the diagram, such as \emph{eyeline}, \emph{rump}, \emph{eyering}, \etc. The list of these words as well as examples, and the morphology diagram were picked based on consultation with an ornithology hobbyist.
The workers were also explicitly instructed to describe only the target image, in an accurate manner, mentioning details that are present in the target image, as opposed to providing jusitifications that talk about features that are absent.

The initial rounds of data collection revealed some interesting corner cases that caused some ambiguity. For example, some workers were confused whether a part of the bird should be called gray or white, because it could appear gray either because the part was white, and in shadow, or the part was actually gray. After these initial rounds of feedback, we proceeded to collect the entire dataset.

\section{Reasoning Speaker Performance Analysis}\label{sec:misc}
\begin{figure}
\includegraphics[width=\columnwidth]{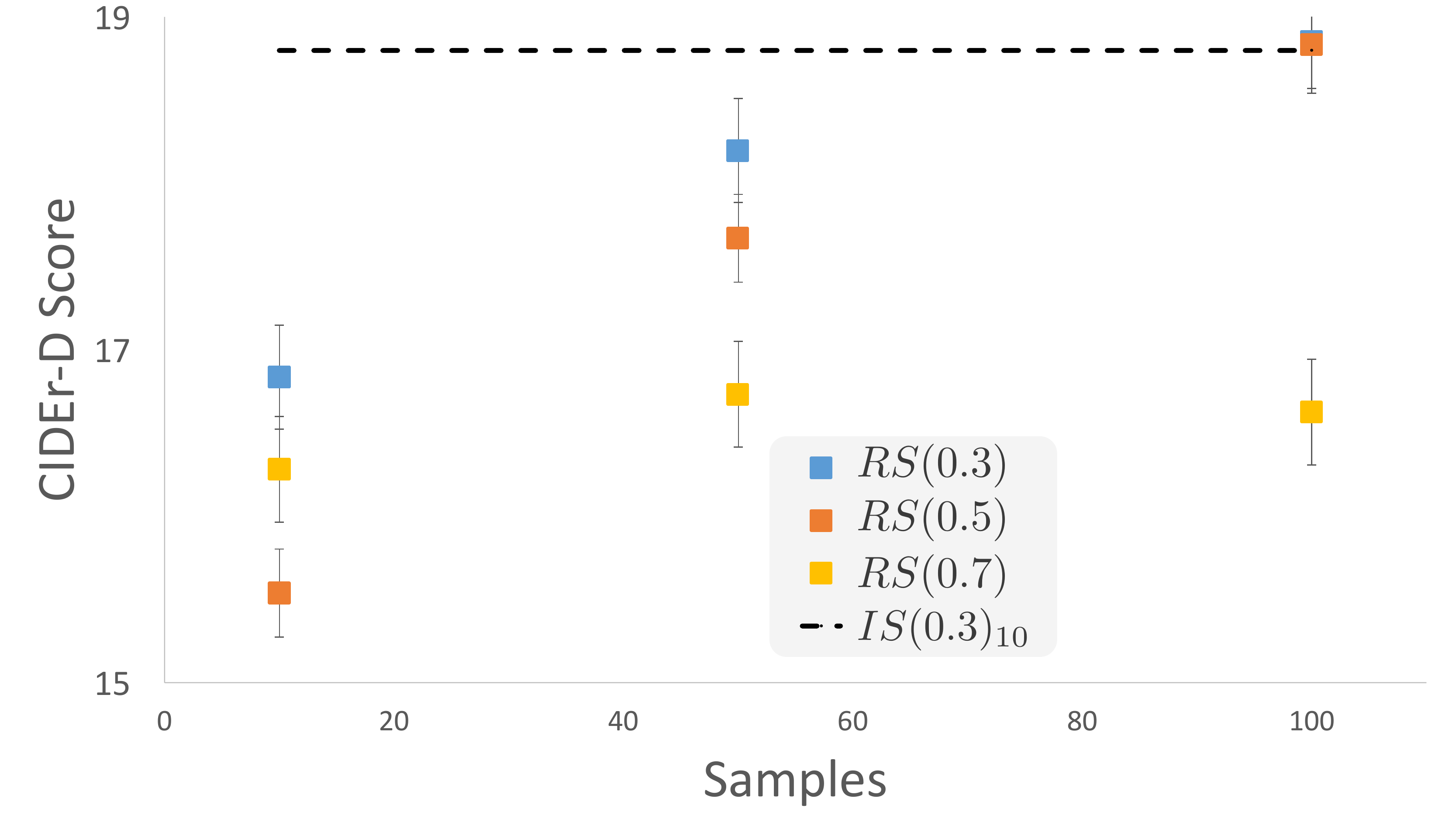}
\caption{We plot how the CIDEr-D score of the \pragma{} baseline (y-axis) varies with the number of samples (x-axis) for different values of $\lambda$. We see that for $\lambda=0.5$, the performance of the \pragma{} method keeps increasing with the number of samples, reaching the performance of our \groundjust{} approach at 100 samples. The \groundjust{} method is shown for reference at a beam size of 10. Thus our approach (\groundjust) is able to give better results for a lower computational cost.}
\label{fig:pragma_plot}
\end{figure}

In this section, we provide more details on how the performance of our adaptation of Andreas, and Klein~\cite{pragma}, namely the \pragma{} approach varies as we sweep over the number of samples we draw from the model for $\lambda=0.3$, $\lambda=0.5$, and $\lambda=0.7$. We note that for $\lambda=0.5$, the \pragma{} approach approaches the best performance from our \groundjust{} approach as we draw 100 samples from the model (\reffig{fig:pragma_plot}). Interestingly, our \groundjust{} model is only evaluated with a beam size of 10. Thus our model is able to perform more efficient search for discriminative sentences than a sampling, and re-ranking based approach like~\pragma{}. It is easy to note that, in case we were willing to spend time to enumerate over all exponentially-many sentences, we would find the optimal solution in worst case exponential time -- most approximate inference techniques in such a setting offer a time vs. optimality tradeoff. Our approach seems to fit this tradeoff better than the~\pragma{} approach based on this empirical evidence.

\section{Comparison to a Trained Listener}\label{sec:trained_listener}
In the main paper we showed comparisons to the \pragma{} baseline with
the same listener model as our approach, which uses the log-likelihood
ratio (\refeq{eqn:introspector}) to assess discriminativeness (and thus needs no further training).
We did this as an apples to apples comparison to systematically
study the benefit of joint inference over the speaker and listener
as opposed to sampling and re-ranking.

For completeness, we report results here using a trained
listener following the architecture choices of~\cite{pragma} for the justification task. We construct reasoning speaker RS($\lambda$) with this trained listener (RS($\lambda$)-TL) and do sampling (sample size 10) and re-ranking, based on log-odds from the listener plugged into Eqn. 1, main paper. As reported in the main paper, \rama{RS($\lambda$)-TL gets
to a best CIDEr-D of 16.2$\pm$0.3 on CUB-Justify validation at $\lambda=0.5$, which is lower than our approach (\emph{semi-blind-}IS($\lambda$) gets to 18.4$\pm$0.2) -- this illustrates the benefit of joint inference.}

For comparsion, we also evaluate against two other reasoning speaker approaches with different rankers (at $\lambda=0$, which only uses the listener, \rama{as opposed to listener+speaker, to directly compare listeners} for ranking): 1) introspector (same as RS(0), main paper), and 2) a chance ranker (RS(0)-R), which randomly scores a class for a sentence. We find that a trained listener (14.7$\pm$0.2) does marginally better than RS(0) (13.9$\pm$0.2) which is in turn better than RS(0)-R (13.4$\pm$0.2). Thus a trained listener does have marginal impact on performance, but the larger factor affecting performance is sampling, which our \emph{semi-blind-}IS($\lambda$) approach is able to do more effectively.

{\small
	\bibliographystyle{ieee}
	\bibliography{bib,rama,consensus_cvpr_camera_ready}
}

\end{document}